\documentclass[letterpaper, 10 pt, conference]{ieeeconf}  
\IEEEoverridecommandlockouts
\usepackage{comment}
\usepackage{cite}
\newcommand{\eat}[1]{}
\usepackage{booktabs}
\usepackage{multicol}
\usepackage{multirow}
\usepackage{diagbox}
\newcolumntype{K}[1]{>{\centering\arraybackslash}p{#1}}

\usepackage{color}
\usepackage [autostyle, english = american]{csquotes}
\MakeOuterQuote{"}
\usepackage{xcolor}
\let\labelindent\relax 
\usepackage{enumitem}

\usepackage{mathtools}

\usepackage{amsmath}
\usepackage{amstext}
\usepackage{amssymb}
\usepackage{amsfonts}
\usepackage{float}

\usepackage{amsthm}  
\usepackage[normalem]{ulem} 

\newtheorem{lemma}{Lemma}
\newtheorem{definition}{Definition}

\newtheorem{remark}{Remark}

\newtheorem{assumption}{Assumption}

\usepackage{subcaption}
\usepackage{caption}
\usepackage{todonotes}

\newcommand{\todoall}[1]{\todo[inline,color=red!10!white]{#1}}

\makeatletter

\newcommand{\Rmnum}[1]{\expandafter\@slowromancap\romannumeral #1@}
\usepackage{savesym}

\usepackage{algorithm}
\usepackage{algorithmicx} 
\usepackage{algpseudocode} %
\savesymbol{AND}
\usepackage[group-separator={,},group-minimum-digits={3}]{siunitx}

\usepackage{graphicx} 
\usepackage{epsfig} 

\usepackage{times} 
\usepackage{amsmath} 
\usepackage{amssymb}  
\usepackage{comment}
\makeatletter
\let\NAT@parse\undefined
\makeatother
\usepackage{hyperref}
\hypersetup{
   colorlinks=true,
    linkcolor= blue,
    allcolors=blue,
    citecolor = blue,
    filecolor=black,
    urlcolor=blue,
    }
\usepackage{mathrsfs}

\usepackage[symbol]{footmisc}

\usepackage[protrusion=true,expansion=true]{microtype}

\usepackage[extramath,probability,reviewmark]{mylatexdefs}

\title{\LARGE \bf
{Multi-Robot Cooperative Navigation in Crowds: A Game-Theoretic Learning-Based Model Predictive Control Approach}}
\author{Viet-Anh Le$^{1,2}$, Vaishnav Tadiparthi$^{3*}$, Behdad Chalaki$^{3*}$\\ Hossein Nourkhiz Mahjoub$^3$, Jovin D'sa$^3$, Ehsan Moradi-Pari$^3$, Andreas A. Malikopoulos$^{2,4}$
\thanks{\scriptsize$^*$Both authors contributed equally.}
\thanks{\scriptsize$^1$Department of Mechanical Engineering, University of Delaware, DE 19716 USA (Email: vietale@udel.edu).
\scriptsize$^2$System Engineering, Cornell University, NY 14850 USA.
This work was conducted during V.-A. Le's internship at Honda Research Institute.}
\thanks{\scriptsize$^3$Honda Research Institute USA, Inc. (Email: \{vaishnav\_tadiparthi; behdad\_chalaki;  hossein\_nourkhizmahjoub; jovin\_dsa; emoradipari \}@honda-ri.com).}
\thanks{\scriptsize$^4$School of Civil and Environmental Engineering, Cornell University, NY 14850 USA (Email: amaliko@cornell.edu).}
}
\begin{document}

\maketitle
\thispagestyle{empty}
\pagestyle{empty}

\begin{abstract} 
In this paper, we develop a control framework for the coordination of multiple robots as they navigate through crowded environments. 
Our framework comprises of a local model predictive control (MPC) for each robot and a social long short-term memory model that forecasts pedestrians' trajectories. 
We formulate the local MPC formulation for each individual robot that includes both individual and shared objectives, in which the latter encourages the emergence of coordination among robots. 
Next, we consider the multi-robot navigation and human-robot interaction, respectively, as a potential game and a two-player game, then employ an iterative best response approach to solve the resulting optimization problems in a centralized and distributed fashion. 
Finally, we demonstrate the effectiveness of coordination among robots in simulated crowd navigation.

\end{abstract}
\section{Introduction}

Social crowd navigation for mobile robots among human pedestrians remains a challenging task due to complex human-robot and human-human interactions, as well as the uncertainty and unpredictability of human behavior.
Moreover, robots' behavior in crowd navigation is not only required to be safe, robust, and efficient like other robotic applications but also needs to ensure social compatibility in motion with human pedestrians.
This is why social crowd navigation has attracted considerable attention in recent years \cite{mavrogiannis2023core}. 

There have been several applications where a team \cite{Malikopoulos2021} of autonomous systems must operate while interacting with humans, such as connected and automated driving in mixed traffic \cite{chalaki2023minimally}, robotic assistance devices \cite{su2021toward}, \etc 
Multi-robot navigation in crowds is a typical application of such cyber-physical-human systems \cite{alleyne2023control} that is significantly more challenging than single robots, as it involves multi-agent coordination and control for human-robot interaction.
In this paper, we aim to consider and address the problem of multi-robot cooperative navigation in crowds.

\subsection{Related Work}

The navigation problem for robots in crowds can be approached through two main strategies: \textit{decoupled prediction and planning} and \textit{coupled prediction and planning} \cite{mavrogiannis2023core}.
In the decoupled approach, human agents are considered dynamic obstacles and thus, mobility-based interactions are ignored.
As a result, ``\textit{freezing robot problem}'' and ``\textit{reciprocal dance problem}'' can occur \cite{mavrogiannis2023core} and cause discomfort to pedestrians nearby.
In contrast, mutual interactions between human and robot agents are embedded in the coupled approach by appropriately incorporating human motion prediction into the problem of robot navigation.
This approach has been observed somewhat to mitigate the freezing robot and reciprocal dance problems, thereby making it a recent trend in crowd navigation research.

Several recent studies in crowd navigation have developed reinforcement learning (RL) methods in which a control policy for a robot is approximated by deep neural networks, \eg collision avoidance deep RL
(CADRL) \cite{chen2017socially}, long short-term memory (LSTM)-RL \cite{everett2018motion} , socially aware RL (SARL) \cite{chen2019crowd}, as well as graph-based techniques \cite{liu2023intention}. 
However, the efficacy of RL policies is often degraded if the robot encounters new scenarios that are different from the training scenario.

On the other hand, optimization-based approaches such as model predictive control (MPC) do not depend on offline training and can generalize well. 
In MPC, we optimize the trajectories
of robots over a finite control horizon given certain prediction models of human trajectories.
Various human prediction models have been used together with MPC in the literature, such as constant velocity \cite{brito2021go,akhtyamov2023social}, Kalman filters
\cite{vulcano2022safe}, intention-enhanced optimal reciprocal collision avoidance \cite{chen2021interactive},
social generative adversarial networks (GAN)
 \cite{poddar2023crowd}, and LSTM \cite{lindemann2023safe}.
Due to the complexity of human behavior, recent data-driven methods such as 
Social-LSTM \cite{alahi2016social}, Social-GAN \cite{gupta2018social}, Social-NCE \cite{liu2021social}, sparse Gaussian processes \cite{trautman2017sparse} have demonstrated more accurate trajectory prediction compared to domain knowledge-based models.
Thus, combining machine learning techniques with MPC 
has the potential to enhance the efficacy of overall navigation. 

\subsection{Contributions of the paper}

While there are increasing studies on social navigation for an individual robot in crowds, the problem of \emph{multi-robot cooperative navigation} among human pedestrians has not been fully explored.
In this paper, we propose an interaction-aware control framework to coordinate robots among human pedestrians using \emph{game-theoretic MPC} and \emph{deep learning-based human trajectory prediction}.
To the best of our knowledge, the most relevant study to our work in the literature is \cite{zhu2021learning}, which focused on multi-robot collision avoidance while the humans were modeled as dynamic obstacles with constant velocities. Thus, the human-robot interaction was not considered.
We first formulate an MPC problem to find the optimal control actions for the robots in a receding horizon fashion and integrate Social-LSTM, a state-of-the-art trajectory prediction model, into the MPC formulation.
To solve the resulting problem, we utilize the concept of the potential game \cite{la2016potential} for multi-robot coordination combined with a two-player game for human-robot interaction to develop two algorithms for centralized and distributed MPC.
We demonstrate the effectiveness of our proposed framework through numerical simulations of a crowd navigation scenario, wherein the robots exhibit coordinated flocking behavior while simultaneously navigating toward their respective goals.

\subsection{Organization of the paper}

The rest of the paper is organized as follows.
In Section~\ref{sec:problem}, we present our problem statement for multi-robot navigation in crowds.
We formulate the MPC framework with the Social-LSTM human trajectory model in Sections~\ref{sec:mpc} and \ref{sec:example}.
In Section~\ref{sec:alg}, we present two algorithms to solve the problem in centralized and distributed manners.
We show the simulation results and analysis in Section~\ref{sec:sim} and draw some conclusions in Section~\ref{sec:conclu}.

\section{Problem Statement}
\label{sec:problem}

\begin{figure}
\centering
\includegraphics[width=0.57\linewidth, bb = 220 135 570 475, clip=true]{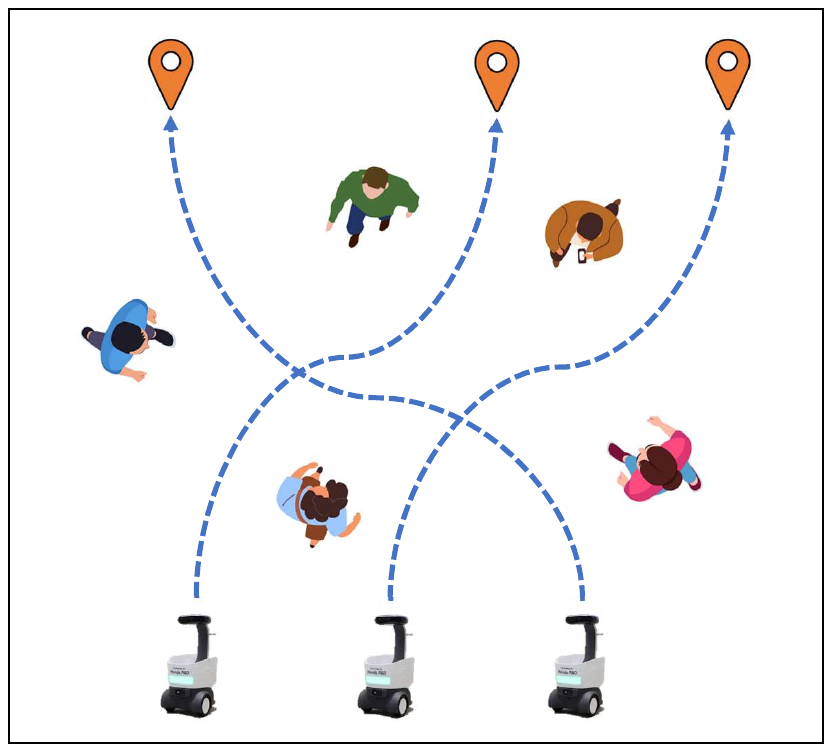}
\caption{An example of multi-robot navigation in a crowd.}
\label{fig:example}
\vspace{-7mm}
\end{figure}

We consider an environment $\WWW \subset \RR^2$ with $N\in\NN$ agents including $M\in\NN$ robots and $N-M$ human pedestrians, illustrated in Fig.~\ref{fig:example}.
We denote $\RRR = \{ 1, \dots, M \}$, $\HHH = \{ N-M+1, \dots, N \}$, and $\RRR\HHH=\RRR\cup\HHH$ the sets of robots, pedestrians, and all agents respectively.

At time step $k\in\NN$, let ${\bb{s}_{i,k} = [s^x_{i,k}, s^y_{i,k}]^\top \in \WWW}$, ${\bb{v}_{i,k} = [v^x_{i,k}, v^y_{i}]^\top \in \RR^2}$, and ${\bb{a}_{i,k} = [a^x_{i,k}, a^y_{i,k}]^\top \in \RR^2}$ be the vectors corresponding to position, velocity, and acceleration of the robot $i\in\RRR$ in Cartesian coordinates, respectively, where each vector consists of two components for $x-$ and $y-$ axis. 
Let $\bb{x}_{i,k} = [\bb{s}_{i,k}^\top, \bb{v}_{i,k}^\top]^\top$ and $\bb{u}_{i,k} = \bb{a}_{i,k}$ be the vectors of states and control actions for the robot $i$ at time step $k$, respectively, while $\bb{s}_{j,k} \in \WWW$ denotes the position of pedestrian $j\in\HHH$ at time step $k$. For simplicity in notation, we use $X_{i,t_1:t_2}$ for agent $i\in\RRR\HHH$ to denote the concatenated arbitrary vector $X$ from time step $t_1$ to time step $t_2$. 
We denote the state and action spaces for robot $i$ as $\XXX_i$ and $\UUU_i$, \ie $\bb{x}_{i,k} \in \XXX_i$ and $\bb{u}_{i,k} \in \UUU_i$, respectively.

\begin{assumption}
\label{assp:set}
We assume that $\XXX_i$ and $\UUU_i$  are nonempty, compact, and connected sets for all $i \in \RRR$.
\end{assumption}

The goal of social multi-robot navigation is to navigate each robot $i$ from the initial positions $\bb{s}_{i}^\mathrm{init}\in\WWW$ to their goals $\bb{s}_{i}^\mathrm{goal}\in\WWW$ while interacting with the human pedestrians to avoid collisions and minimizing any discomfort during motion.
We consider the following assumption to facilitate positioning and communication between the robots for coordination.

\begin{assumption}
\label{assp:positioning}
The robots' and pedestrians’ real-time positions can be obtained by a positioning system and stored by a central coordinator (\eg a computer). 
The robots and the coordinator can exchange information through wireless communication.
\end{assumption}


\section{Model Predictive Control with Social-LSTM}
\label{sec:mpc}

In our proposed framework, we combine a Social-LSTM model, which 
predicts the future trajectories of human pedestrians with an MPC formulation to derive the optimal control actions for the robots.
Next, we formally define the Social-LSTM model and MPC formulation in detail.

\subsection{Human Motion Prediction using Social-LSTM}

Let $t\in\NN$ be the current time step, $H\in \mathbb{Z}^+$ be the control/prediction horizon length, and $\III_t = \{t,t+1 \dots, t+H-1 \}$ be the set of time steps in the next control horizon.
The human prediction model aims at predicting the trajectories of pedestrians over a prediction horizon of length $H$ given the observations over $L\in \mathbb{Z}^+$ previous time-steps of all agents' trajectories.
Social-LSTM was developed in \cite{alahi2016social} for jointly predicting the multi-step trajectory of multiple agents. 
Inspired by \cite{gupta2023interaction}, we consider recursive prediction for the pedestrians' positions using the single-step Social-LSTM model denoted by $\phi(\cdot):\mathbb{R}^{2|\RRR\HHH|L}\rightarrow\mathbb{R}^{2|\HHH|}$ as follows 
\begin{equation}
\label{eq:lstm_recursive}
\bb{s}_{\HHH,k+1} = \phi (\bb{s}_{\RRR\HHH,k-L+1:k}),\, \forall k \in \III_t.
\end{equation}
Note that we have made several modifications to the prediction model \eqref{eq:lstm_recursive} compared to the original Social-LSTM model proposed in the work of \cite{alahi2016social}.
First, the Social-LSTM model in \cite{alahi2016social} performs simultaneous prediction for all $N$ agents, \ie both robots and humans. However, in \eqref{eq:lstm_recursive}, we precisely extract the positions of human pedestrians from the joint predictions.
Additionally, we consider the Social-LSTM model with single-step prediction, where predictions are made recursively throughout the control horizon at each time step instead of a multi-step prediction for the entire control horizon.
The recursive technique yields a coupled prediction and planning approach while using the multi-step prediction model leads to a decoupled prediction and planning approach, \ie ignoring the mutual dependence between humans' and robots' future behavior.
Last, at each time-step of the control horizon, the humans' predicted positions in previous time-steps are used as the inputs of the Social-LSTM model, while the predicted positions of the robots are discarded since the future robots' positions depend on the solution of the MPC problem.

\begin{remark}
Though we utilize Social-LSTM \cite{alahi2016social} as the human motion prediction model, other deep learning models such as \cite{gupta2018social}, \cite{liu2021social}, \cite{kothari2021human} can be used alternatively.
\end{remark}

\subsection{Model Predictive Control Formulation}

Henceforth, for ease of notation, we omit the time subscript of the variables if it refers to all time-steps within the control horizon.
For example, we use $\bb{u}_i$, $\bb{x}_i$, and $\bb{s}_j$, $\forall i \in \RRR, \, j \in \HHH$, instead of $\bb{u}_{i,t:t+H-1}$, $\bb{x}_{i,t+1:t+H}$, and $\bb{s}_{j,t+1:t+H}$.
We consider that the dynamics of each robot are governed by a discrete-time double-integrator model:
\begin{equation}
\label{eq:dynamics}
\begin{split}
\bb{s}_{i,k+1} &= \bb{s}_{i,k} + \tau \bb{v}_{i,k} + \frac{1}{2} \tau^2 \bb{a}_{i,k}, \\
\bb{v}_{i,k+1} &= \bb{v}_{i,k} + \tau \bb{a}_{i,k},
\end{split}
\end{equation}
where $\tau \in \RRplus$ is the sampling time period. 
Each robot has an individual objective $J_i (\bb{u}_i, \bb{x}_i, \bb{s}_{\HHH})$ and a shared objective with other robots $J_{-i} (\bb{x}_i, \bb{x}_{-i})$, where we define $\bb{x}_{-i}^\top = [\bb{x}_{j}^\top]_{\forall j \in \RRR, j \neq i}$ as the concatenation of the state vectors of all other robots in the set $\RRR$.
We consider pairwise shared objectives between robots
\begin{equation}
J_{-i} (\bb{x}_i, \bb{x}_{-i}) = \sum_{j \in \RRR \setminus \{i\}} J_{ij}(\bb{x}_i, \bb{x}_j).
\end{equation}
We formulate the local MPC problem for robot $i$ at time $t$:
\begin{equation}
\label{eq:agent_mpc}
\begin{split}
&\minimize_{\bb{u}_i}
\; J_i (\bb{u}_i, \bb{x}_i, \bb{{s}}_{\HHH}) + \sum_{j \in \RRR \setminus \{i\}} J_{ij} (\bb{x}_i, \bb{x}_{j}),
\\ 
& \subjectto 
\; \eqref{eq:lstm_recursive}, \, \eqref{eq:dynamics}, \bb{u}_{i,k} \in \UUU_i, \bb{x}_{i,k+1} \in \XXX_i, \, \forall k \in \III_t,
\\
& \text{given:} 
\; \bb{s}_{\RRR\HHH,t-L+1:t}, \bb{x}_{i,t}.
\end{split}
\end{equation}
Note that any constraints, \eg collision avoidance constraints, can be included in the objective function by penalty functions with sufficiently large weights.

\begin{assumption}
\label{assp:func}
We assume that the individual objective $J_i$ and shared objective $J_{ij}$, for all $i,j \in \RRR$, are continuous and differentiable functions everywhere over $\mathcal{X}$ and $\mathcal{U}$.
\end{assumption}


\section{Multi-robot Cooperative Navigation}
\label{sec:example}

In this section, we illustrate the MPC formulation \eqref{eq:agent_mpc} by considering the scenario illustrated in Fig.~\ref{fig:example}, where multiple robots navigate to their goals among several human pedestrians.
The robots need to reach their goals and avoid collision with human pedestrians and other robots.
Moreover, since the robots move in the same direction, they can coordinate for moving in a flock while navigating to the goals.
We assume that the states and control inputs of robots $i\in\RRR$ are subjected to the following bound constraints: 
\begin{equation}
\!\!\! -v_{\max} \le v^x_{i,k}, v^y_{i,k} \le v_{\max}, \, 
-a_{\max}  \le a^x_{i,k}, a^y_{i,k} \le a_{\max},
\label{eq:bound_a}
\end{equation}
where $v_{\max} \in \RRplus$ and $a_{\max} \in \RRplus$ are the maximum speed and acceleration of the robots, respectively.

We formulate individual and shared objectives by weighted sums of several features.
For robot $i$, we consider a tracking minimization to a desired reference trajectory 
\begin{equation}
\! J_i^{\mathrm{goal}} (\bb{s}_i) \!=\! \frac{\sum_{k=t}^{t+H-1} (\bb{s}_{i,k+1} - \bb{s}_{i,k+1}^{\mathrm{ref}})^\top 
(\bb{s}_{i,k+1} - \bb{s}_{i,k+1}^{\mathrm{ref}})}{\norm{\bb{s}_{i,t} - \bb{s}_{i}^\mathrm{goal}}_2 + \delta}, 
\end{equation}
where $\bb{s}_{i,k+1}^{\mathrm{ref}}$ is the desired position at time $k+1$.
The feature is normalized 
by the distance to the goal 
where $\delta \in \RRplus$ is a small positive number. 
Consequently, the significance of the goal-reaching objective increases as the robot approaches its intended destination. We compute the desired reference trajectory for robot $i$ based on a straight line to its goal
\begin{equation}
\!\! \bb{s}_{i,k+1}^{\mathrm{ref}} \! = \! \bb{s}_{i,k}^{\mathrm{ref}} 
\!+\! \min \! \Big \{ \! \tau v_{\max}, \! \norm{\bb{s}_i^\mathrm{goal} - \bb{s}^{\mathrm{ref}}_{i,k}} \! \Big \} \!   
\frac{\bb{s}_i^\mathrm{goal} - \bb{s}^{\mathrm{ref}}_{i,t}}{\norm{\bb{s}_i^\mathrm{goal} - \bb{s}^{\mathrm{ref}}_{i,t}}}, 
\end{equation}
for $k\in\III_t$ and $\bb{s}^{\mathrm{ref}}_{0,t} = \bb{s}_{0,t}$.
In addition, we minimize the acceleration and jerk 
of the robot’s motion by: 
\begin{align}
J_i^{\mathrm{acce}} (\bb{u}_i) &= \sum_{k=t}^{t+H-1} \bb{u}_{i,k}^\top \bb{u}_{i,k}, \label{eq:J_acce}\\    
J_i^{\mathrm{jerk}} (\bb{u}_i) &= \sum_{k=t}^{t+H-1} ( \bb{u}_{i,k} - \bb{u}_{i,k-1} )^\top ( \bb{u}_{i,k} - \bb{u}_{i,k-1} ). \label{eq:J_jerk}
\end{align}
To prevent collision between a robot and any human pedestrian, we impose the following constraint that the distance between the robot and each pedestrian is greater than a safe speed-dependent distance
\begin{equation}
\label{eq:human_collision}
\begin{multlined}
\!\! g_{ij}(\bb{x}_{i,k+1}, \bb{s}_{j,k+1}) = d_{\min}^2 + \rho \norm{\bb{v}_{i,k+1}}_2^2 \\ \quad
- \norm{\bb{s}_{i,k+1} - \bb{s}_{j,k+1}}_2^2 \le 0,
\end{multlined}
\end{equation}
$\forall i \in \RRR, j \in \HHH$,
where $d_{\min} \in \RRplus$ is a minimum allowed distance and $\rho \in \RRplus$ is a scaling factor.
We include a speed-dependent term in the above constraint to ensure that the robot stays farther away from the humans while moving at high speed, leading to less human discomfort.
Similarly, we consider an inter-robot collision avoidance constraint 
\begin{equation}
\label{eq:robot_collision}
\! g_{ij} (\bb{x}_{i,k+1}, \bb{x}_{j,k+1}) = d_{\min}^2 - \norm{\bb{s}_{i,k+1} - \bb{s}_{j,k+1}}_2^2 \le 0, \\
\end{equation}
$\forall\, i,j \in \RRR$.
The omission of a speed-dependent term in \eqref{eq:robot_collision} is justified by the desire to maintain consistency with the subsequent shared objective, encouraging robots to move towards their goals while staying close to one another. 

We incorporate the collision avoidance constraint into the objective function as a soft constraint by using a smoothed max penalty function as follows
\begin{equation}
J_{ij}^{\mathrm{coll}} (\bb{x}_{i}, \bb{x}_{j}) 
= \sum_{k=t}^{t+H-1} \mathrm{smax} \big( g_{ij}(\bb{x}_{i,k+1}, \bb{x}_{j,k+1}) \big).
\end{equation}
The smoothed max penalty function is defined as
$\mathrm{smax}(z) = \frac{1}{\mu} \log \big( \exp(\mu z) + 1 \big),$ 
where $\mu \in \RRplus$ is a parameter that adjusts the smoothness of the penalty function.
Additionally, the shared objective between robots $i$ and $j$ includes the following flocking control objective: 
\begin{equation}
\label{eq:Jijf}
J_{ij}^{\mathrm{floc}} (\bb{x}_{i}, \bb{x}_{j}) = \sum_{k=t}^{t+H-1} \big( \norm{\bb{s}_{i,k+1} - \bb{s}_{j,k+1}}_2 - d_{ij} \big)^2, 
\end{equation}
where $d_{ij} \in \RRplus$ is the desired distance between two robots while moving in a flock.
For everywhere differentiability, 
we approximate the $L_2$-norm by $\norm{\bb{z}}_2 \approx \sqrt{\bb{z}^\top \bb{z} + \delta}$,
where $\delta \in \RRplus$ is a small positive number.

Hence, the individual objective can be given by: 
\begin{equation}
\label{eq:Ji}
\begin{multlined}
J_i (\bb{u}_i, \bb{x}_i, \bb{s}_{\HHH})  
= \omega_i^{\mathrm{goal}} J_i^{\mathrm{goal}} (\bb{x}_i) + \omega_i^{\mathrm{acce}} J_i^{\mathrm{acce}} (\bb{u}_i) \\
+ \omega_i^{\mathrm{jerk}} J_i^{\mathrm{jerk}} (\bb{u}_i) + \sum_{j \in \HHH} \omega_{ij}^{\mathrm{coll}} J_{ij}^{\mathrm{coll}} (\bb{x}_i, \bb{s}_j),
\end{multlined}
\end{equation}
and the shared objective is given as follows,
\begin{equation}
\label{eq:Jij}
J_{ij} ( \bb{x}_i, \bb{x}_{j})  
= \omega^{\mathrm{coll}}_{ij} J_{ij}^{\mathrm{coll}} (\bb{x}_i, \bb{x}_j) 
+ \omega^{\mathrm{floc}}_{ij} J_{ij}^{\mathrm{floc}} (\bb{x}_i, \bb{x}_j).
\end{equation}

\section{Centralized vs Distributed Model Predictive Control}
\label{sec:alg}

Game theory is an effective tool for formulating and analysing different types of interactions among agents.
As a result, game-theoretic planning and control have been used widely in multi-agent coordination \cite{wang2020multi} and human-robot interaction \cite{fisac2019hierarchical,liu2023potential}.
In this section, we first recast the problem \eqref{eq:agent_mpc} as a potential game between the robots combined with a two-player game between the robots and the humans.
Then, we derive centralized and distributed MPC algorithms based on an iterative best-response approach.
\subsection{Multi Robot Coordination as a Potential Game}

Given the MPC problem \eqref{eq:agent_mpc} for each robot, we define a game for multi-robot navigation as follows. 

\begin{definition}
\label{def:game}

At each time-step $t$ for control/prediction horizon length $\III_t$, we define an $M$-player game for multi-robot navigation as
$\GGG_{t}^{\RRR} :=  \big \{ \RRR, \mathbb{U}_{\RRR}, 
\{C_{i,t} 
\}_{i \in \RRR} 
\big \}$, where $\mathbb{U}_\RRR := \mathcal{U}_1^H \times \mathcal{U}_2^H \times \dots \times \mathcal{U}_M^H$ 
while $\mathcal{U}_i^H$ is the set of {all} action sequences of robot $i$ in $\III_t$ over a horizon $H$ (also referred to as a strategy set). 
Let $\bb{u}_{i} \in \mathcal{U}_i^H$ be a valid strategy for robot $i$ and a joint strategy adopted by other players be denoted by $\bb{u}_{-i}:= [\bb{u}_j^\top]^\top_{j \in \RRR\setminus \{i\}} \in \mathcal{U}_{-i}^H $.
Then, the cumulative cost for each robot $i$ over $\III_t$ is denoted by $C_{i,t} (\bb{u}_i, \bb{u}_{-i}) := J_i (\cdot) + J_{-i} (\cdot)$ {, which can be evaluated by the objective function in \eqref{eq:agent_mpc}.} 
\end{definition}

The following definition of a continuous exact potential game was presented in \cite{liu2023potential,la2016potential}.

\begin{definition}
If 
for every robot $i\in\RRR$, $\bb{u}_i$ is a nonempty, compact, and connected set ({extending} Assumption~\ref{assp:set}), $C_{i,t} (\cdot)$ is a continuous and differentiable function (Assumption~\ref{assp:func}), then the game $\GGG_{t}^{\RRR}$ is a continuous exact potential game if and only if there exists a function $F_t: \mathbb{U}_\RRR
\rightarrow \RR$ that is continuous and differentiable on $\mathbb{U}_\RRR$ 
and satisfies 
\begin{equation}
\! \frac{\partial C_{i,t} (\bb{u}_i, \bb{u}_{-i})}{\partial \bb{u}_i} 
= \frac{\partial F_t (\bb{u}_i, \bb{u}_{-i})}{\partial \bb{u}_i}, \forall \bb{u}_i \in \mathcal{U}_i^H, \bb{u}_{-i} \in \mathcal{U}_{-i}^H,
\end{equation}
where $\mathcal{U}_i^H$ and $\mathcal{U}_{-i}^H$ are the strategy sets as defined previously.
The function $F_t$ is called a \textbf{potential function} of the game.

\end{definition}

\begin{remark}
The conditions in Definition~\ref{def:game} hold for multi-robot cooperative navigation presented in Section~\ref{sec:example} as given in \eqref{eq:bound_a}, the domains $\UUU_i$ and $\XXX_i$ are nonempty, compact and connected, while the individual and shared objective functions shown in \eqref{eq:Ji} and \eqref{eq:Jij} are everywhere continuous and differentiable. 
\end{remark}

\begin{lemma}
\label{lem:potential}
$\GGG_{t}^{\RRR}$ is a continuous exact potential game if $J_{ij} (\bb{x}_i, \bb{x}_j) = J_{ji} (\bb{x}_j, \bb{x}_i)$ and the potential function at time step $t$ is computed by
\begin{equation}
\label{eq:poten}
\! F_t(\bb{u}_\RRR, \bb{{s}}_{\HHH}) = \! \sum_{i \in \RRR} J_i (\bb{u}_i, \bb{x}_i, \bb{{s}}_{\HHH}) + \!\!\! \sum_{i,j \in \mathcal{R}, i\neq j} \!\! J_{ij} (\bb{x}_i, \bb{x}_{j}),
\end{equation}
where given the system dynamics \eqref{eq:dynamics}, the right-hand side of \eqref{eq:poten} can be expressed as a function of the joint strategy $\bb{u}_\RRR$ and the pedestrians' predicted trajectory $\bb{s}_\HHH$ in the left-hand side.
\end{lemma}

The proof for Lemma~\ref{lem:potential} directly follows from the proof of Theorem 8 in \cite{liu2023potential}. 
In order to satisfy Lemma~\ref{lem:potential} for the scenario presented in Section~\ref{sec:example},
we choose $\omega_{ij}^\mathrm{coll} = \omega_{ji}^\mathrm{coll}$ and $\omega_{ij}^\mathrm{floc} = \omega_{ji}^\mathrm{floc}$, $\forall i,j \in \mathcal{R}, i\neq j$. 

\subsection{Centralized MPC (CMPC)}

Since a Nash equilibrium can be found by optimizing the potential function of the game at each time step, we can implement the multi-robot navigation algorithm in a centralized manner, \ie using only a central coordinator, by solving the following centralized MPC problem:
\begin{equation}
\label{eq:cen_mpc}
\begin{split}
&\minimize_{\bb{u}_\RRR} \; F_t (\bb{u}_\RRR, \bb{{s}}_{\HHH}),
\\ 
& \subjectto 
\;\eqref{eq:lstm_recursive}, \eqref{eq:dynamics}, \bb{u}_{\RRR} \in \mathbb{U}_{\RRR} , \bb{x}_{i,k+1} \in \XXX_i, \\
& \text{given:}  \;
\bb{s}_{\RRR\HHH,t-L+1:t}, \bb{x}_{\RRR,t},  
\end{split}
\end{equation}
where the constraints hold for all $i \in \RRR$ and $k \in \III_t$.
Due to the complexity of the Social-LSTM network, solving the MPC problem \eqref{eq:cen_mpc} would be computationally intractable.
Therefore, we employ iterative best response (IBR), a commonly used algorithm in game theory, to find a Nash equilibrium. 
We first define a two-player game for human-robot interaction.

\begin{algorithm}[tb!]
\small
  \caption{IBR-based centralized MPC}
  \label{alg:ibr_cenmpc}
  \label{alg:iterative}
  \begin{algorithmic}[1]
    \Require $t$, $H$, $j_{\mathrm{max}}$, $\xi$, $\bb{u}_{\RRR}^{(0)}$, $\bb{x}_{\RRR}^{(0)}$, $\bb{s}_{\RRR\HHH,t-L+1:t}$.
    
    \For{$j = 1,2,\dots,j_{\mathrm{max}}$}
    \State Predict  $\bb{s}_\HHH^{(j)}$ using \eqref{eq:lstm_recursive} given $\bb{x}_{\RRR}^{(j-1)}$. and $\bb{s}_{\RRR\HHH,t-L+1:t}$. 
    \State Solve \eqref{eq:cen_mpc} 
    given $\bb{s}_{\HHH}^{(j)}$ to obtain $\bb{u}_{\RRR}^{(j)}$ and $\bb{x}_{\RRR}^{(j)}$.
    \State Compute $F_t^{(j)} = F_t (\bb{u}_\RRR^{(j)}, \bb{{s}}_{\HHH}^{(j)})$.
    \If {$\abs{F_t^{(j)} - F_t^{(j-1)}} \le \xi$}
    \State \Return $\bb{u}_{\RRR}^{(j)}$ 
    \EndIf
    \EndFor
    \State \Return $\bb{u}_{\RRR}^{(j_{\mathrm{max}})}$ 
  \end{algorithmic}
\end{algorithm} 
\setlength{\textfloatsep}{0.05cm}

\begin{definition}
\label{def:interaction}
At each time-step $t$, we define a two-player game 
for human-robot interaction 
as $\GGG_t^{\RRR\HHH} := \left\{ \{\RRR, \HHH\}, \{\mathbb{U}_{\RRR}, \mathbb{R}^{2|\HHH| |\III_t|}\}, \{F_t(\bb{u}_\RRR, \bb{s}_\HHH), I_t(\bb{s}_\RRR, \bb{s}_\HHH)\} \right \}$, where 
\begin{equation}
I_t (\bb{s}_\RRR, \bb{s}_\HHH) = \sum_{k \in \III_t} \norm{\bb{s}_{\HHH,k+1} - \phi (\bb{s}_{\RRR\HHH,k-L+1:k})}_2.
\end{equation}
In this game, the group of robots and the group of humans are considered two players; each player's strategy is defined by the trajectories over the next control horizon.
The cost function for robots is the objective function in \eqref{eq:cen_mpc}, 
while the cost function for human pedestrians is computed by the deviation to the predicted trajectories given by \eqref{eq:lstm_recursive}.
In other words, we assume that the best strategy for human pedestrians is to follow the trajectories given by the output of the Social-LSTM model \eqref{eq:lstm_recursive}.
{ The prediction model is expected to encode the fundamentals of pedestrian motion, such as collision avoidance, path following, and comfort. }
\end{definition}

In the IBR approach, a single agent updates its strategy at each iteration based on its best response to the others' strategies.
Therefore, given the human-robot interaction game (Definition~\ref{def:interaction}), we can solve \eqref{eq:cen_mpc} by sequentially computing the Social-LSTM prediction and optimizing the MPC objective function.
All the steps are performed by the coordinator until the difference in the potential function evaluated at two consecutive iterations is smaller than a threshold $\xi \in \RRplus$, or a maximum number of iterations $j_{\max} \in \NN$ is reached.
The details are provided in Algorithm~\ref{alg:ibr_cenmpc}.
If the algorithm converges, the converged point is a Nash equilibrium \cite{williams2018best,espinoza2022deep}. 
However, it does not always converge; in the worst case, it may cycle between strategies.

\begin{table*}
\centering
\caption{Comparison between centralized MPC (CMPC) and distributed MPC (DMPC).}
\label{tab:cenvsdis}
\begin{tabular}{ K{2.8cm} | K{0.8cm} K{0.8cm} | K{0.8cm} K{0.8cm} | K{0.8cm} K{0.8cm} | K{0.8cm} K{0.8cm} | K{0.8cm} K{0.8cm}  }
\toprule[1.2pt]
Numbers of Pedestrians & \multicolumn{2}{c|}{5} & \multicolumn{2}{c|}{6} & \multicolumn{2}{c|}{7} & \multicolumn{2}{c|}{8} & \multicolumn{2}{c}{9} \\ 
\midrule[0.6pt]
\backslashbox{Metrics}{Methods} & CMPC & DMPC & CMPC & DMPC & CMPC & DMPC & CMPC & DMPC & CMPC & DMPC \\
\midrule[0.6pt]
Success rate (\%) 
& $\bf{94.0}$ & $93.4$ 
& $\bf{91.9}$ & $91.8$ 
& $\bf{90.0}$ & $88.7$ 
& ${86.7}$ & ${\bf{86.9}}$ 
& $\bf{84.5}$ & $83.9$ \\
Average travel time (s)  
& $\bf{17.9}$ & $18.1$ 
& $\bf{18.6}$ & $18.7$ 
& $\bf{19.1}$ & $19.2$ 
& $\bf{19.6}$ & $19.8$ 
& $\bf{20.1}$ & $20.3$ \\
Collision rate (\%) 
& $\bf{0.2}$ & $0.3$ 
& ${0.8}$ & $\bf{0.6}$ 
& ${1.3}$ & $\bf{1.1}$ 
& ${1.1}$ & ${1.1}$ 
& $\bf{1.2}$ & $1.9$ \\
Discomfort rate (\%) 
& $\bf{0.5}$ & $0.6$ 
& $\bf{1.2}$ & $1.5$ 
& ${1.3}$ & $\bf{1.1}$ 
& $1.4$ & $\bf{1.3}$ 
& ${2.1}$ & $\bf{1.8}$ \\
\bottomrule[1.2pt]
\end{tabular}
\vspace{-2.5mm}
\end{table*}

\begin{table*}
\centering
\caption{Comparison between centralized MPC with (F) and without (NF) the flocking objective.}
\label{tab:flockvsno}
\begin{tabular}{ K{2.8cm} | K{0.8cm} K{0.8cm} | K{0.8cm} K{0.8cm} | K{0.8cm} K{0.8cm} | K{0.8cm} K{0.8cm} | K{0.8cm} K{0.8cm}  }
\toprule[1.2pt]
Numbers of Pedestrians & \multicolumn{2}{c|}{5} & \multicolumn{2}{c|}{6} & \multicolumn{2}{c|}{7} & \multicolumn{2}{c|}{8} & \multicolumn{2}{c}{9} \\ 
\midrule[0.6pt]
\backslashbox{Metrics}{Methods} & F & NF & F & NF & F & NF & F & NF & F & NF \\
\midrule[0.5pt]
Success rate (\%) 
& $94.0$ & $\bf{96.7}$ 
& $91.9$ & $\bf{94.0}$ 
& $90.0$ & $\bf{93.8}$ 
& $86.6$ & $\bf{88.8}$ 
& $84.7$ & $\bf{86.1}$ \\
Average travel time (s)  
& $17.9$ & $\bf{17.6}$ 
& $18.6$ & $\bf{18.4}$ 
& $19.1$ & $\bf{18.8}$ 
& $19.6$ & $\bf{19.3}$ 
& $20.1$ & $\bf{19.8}$ \\
Collision rate (\%) 
& ${0.2}$ & $0.2$ 
& $\bf{0.8}$ & $0.9$ 
& ${1.3}$ & $\bf{0.4}$ 
& $\bf{1.1}$ & $1.8$ 
& $\bf{1.2}$ & $2.4$ \\
Discomfort rate (\%) 
& $\bf{0.5}$ & $1.4$ 
& $\bf{1.2}$ & $2.2$ 
& $\bf{1.3}$ & $2.0$ 
& $\bf{1.4}$ & $2.2$ 
& $\bf{2.1}$ & ${2.6}$ \\
\bottomrule[1.2pt]
\end{tabular}
\vspace{-4.4mm}
\end{table*}

\begin{figure*} 
\centering
\begin{subfigure}{.19\textwidth} \hspace{-10pt}
\centering
\includegraphics[width=1.05\textwidth]{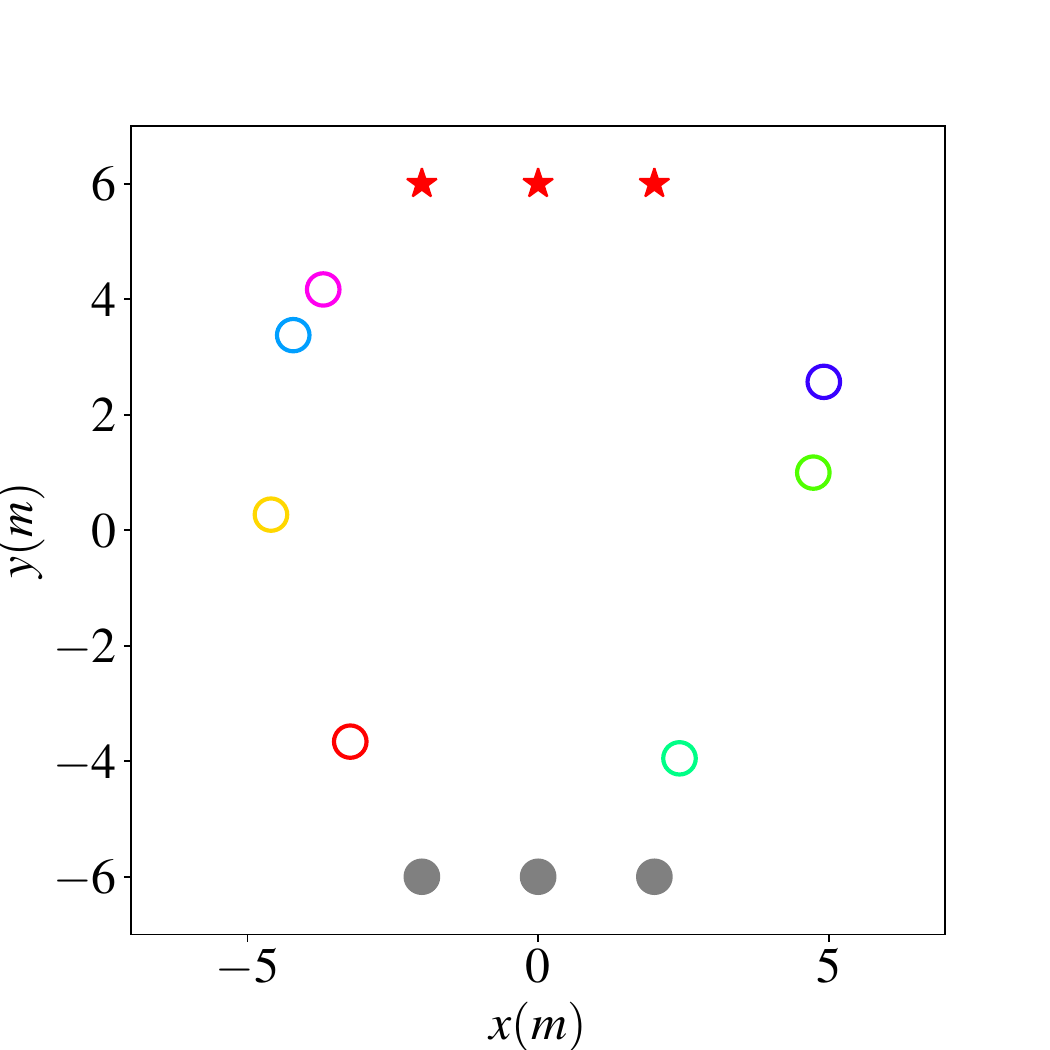}
\caption{$t = \SI{0}{s}$}
\end{subfigure} 
\begin{subfigure}{.19\textwidth} \hspace{-10pt}
\centering
\includegraphics[width=1.05\textwidth]{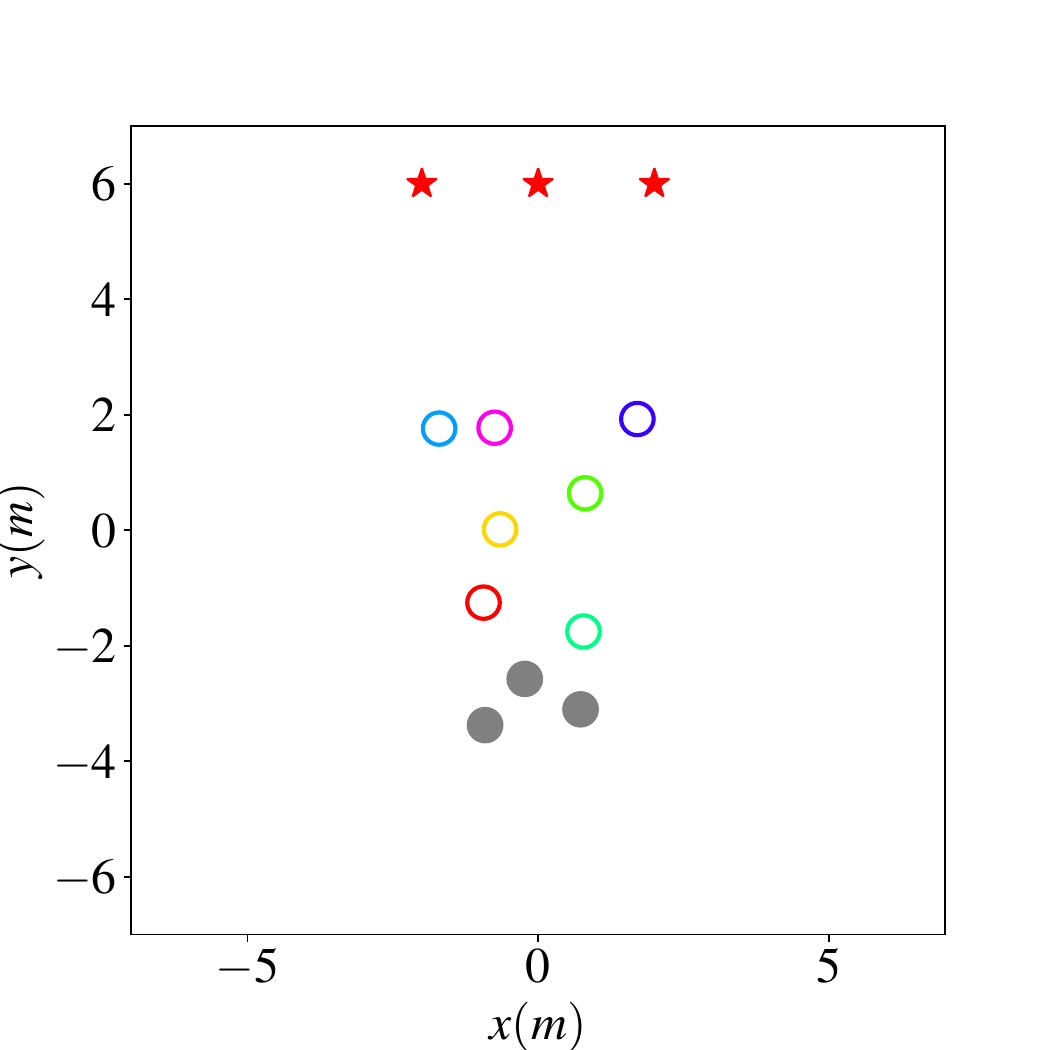}
\caption{$t = \SI{4}{s}$}
\end{subfigure} 
\begin{subfigure}{0.19\textwidth} \hspace{-10pt}
\centering
\includegraphics[width=1.05\textwidth]{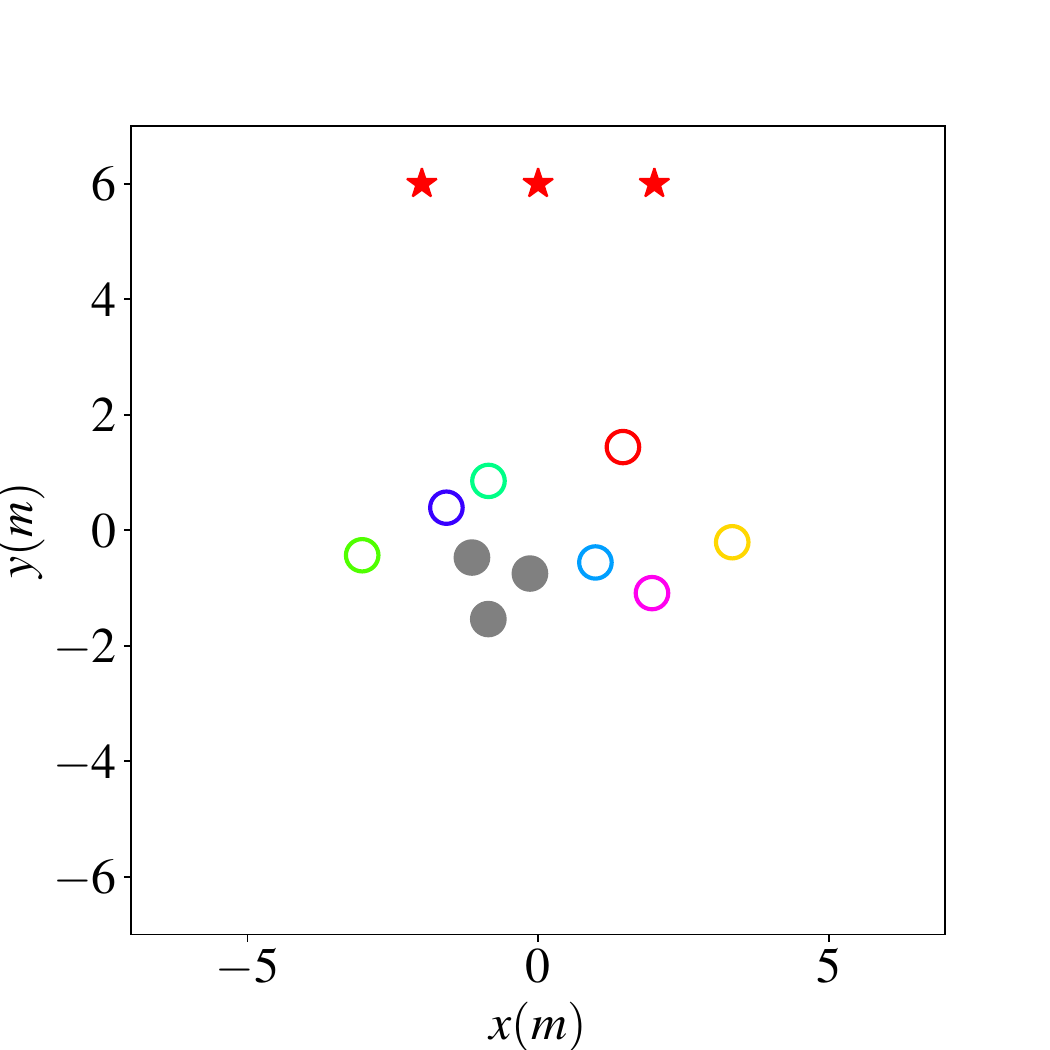}
\caption{$t = \SI{8}{s}$}
\end{subfigure} 
\begin{subfigure}{.19\textwidth} \hspace{-10pt}
\centering
\includegraphics[width=1.05\textwidth]{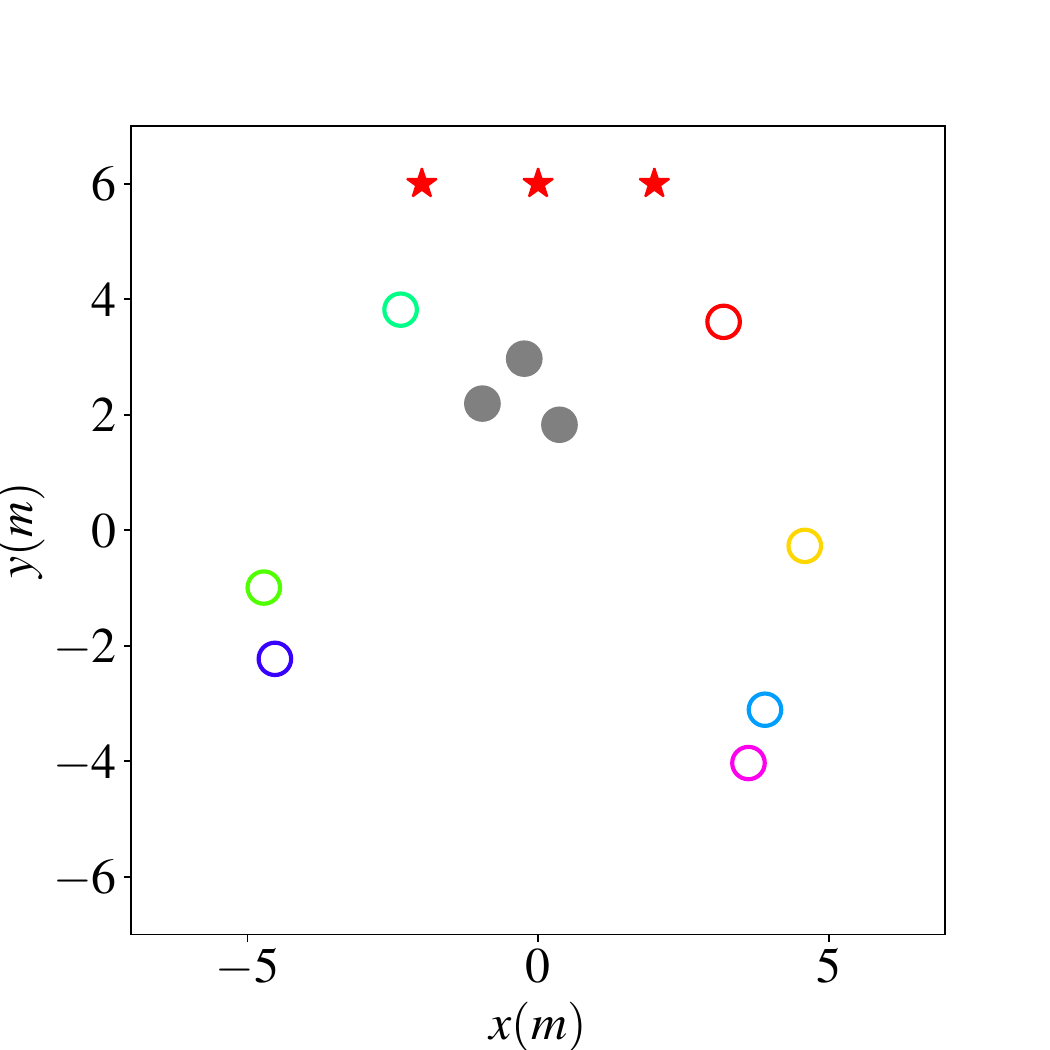}
\caption{$t = \SI{12}{s}$}
\end{subfigure} 
\begin{subfigure}{.19\textwidth} \hspace{-10pt}
\centering
\includegraphics[width=1.05\textwidth]{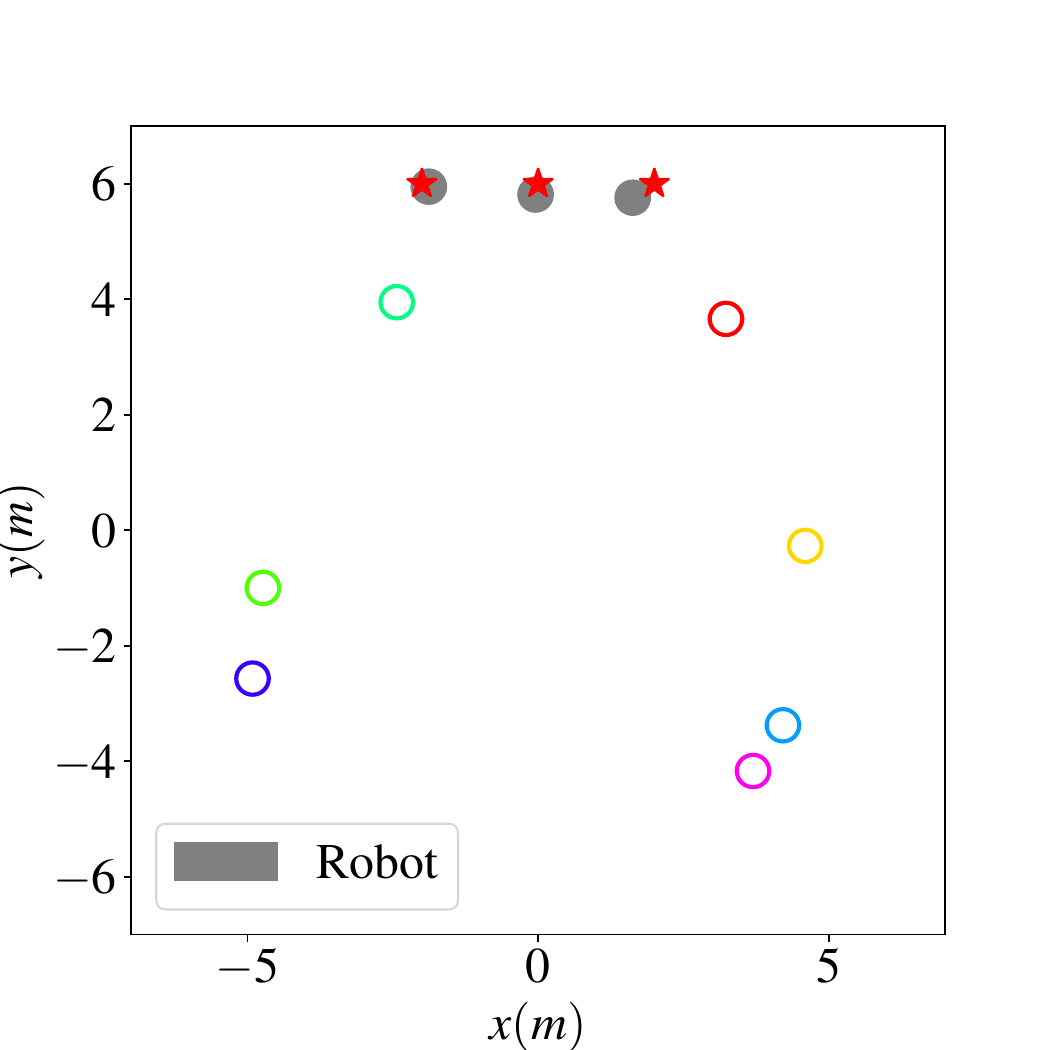}
\caption{$t = \SI{16.4}{s}$}
\end{subfigure} 


\caption{Positions of the robots and human pedestrians at several time steps in a circular crossing simulation. Humans' and robots' positions are marked by unfilled and solid circles, respectively, while red stars represent robots' destinations.}
\label{fig:trajcircle}
\vspace{-5mm}
\end{figure*}

\subsection{Distributed MPC (DMPC)}
We also developed an IBR-based distributed MPC algorithm in which the coordinator and the robots collaborate and communicate to solve the problem.
Each robot repeatedly solves its local MPC problem \eqref{eq:agent_mpc} given the current best-response trajectories of other robots in parallel. 
The robots' current best-response trajectories can be exchanged with the other robots through communication between the coordinator and the robots.
Though the IBR approach advances towards a Nash equilibrium in a continuous exact potential game, convergence may not be achieved within a finite number of iterations \cite{liu2023potential}. 
Hence, to improve the convergence of the IBR algorithm, we can modify the algorithm based on the concept of $\epsilon$-Nash equilibrium \cite{roughgarden2010algorithmic}, defined as follows:

\begin{definition}
A strategy $(\bb{u}_1^*,\bb{u}_2^*,\dots,\bb{u}_M^*)$ is an $\epsilon$-Nash equilibrium if and only if
$\exists\, \epsilon \in \RRplus$ such that $\forall i \in \RRR$; 
\begin{equation}
C_{i,t} (\bb{u}_{i}^*, \bb{u}_{-i}^*)    
\le C_{i,t} (\bb{u}_{i}', \bb{u}_{-i}^*) + \epsilon,\, \forall \bb{u}_{i}' \in \mathcal{U}_i^H.
\end{equation}
\end{definition}
Combining the IBR approach for the multi-robot navigation game and the human-robot interaction game, we derive Algorithm~\ref{alg:ibrdmpc} for solving the distributed MPC problem. 
It starts with an initial guess $\bb{x}_{\RRR}^{(0)}$ of the best-response trajectories for the robots.
At each iteration $j$, the coordinator predicts $\bb{s}_{\HHH}^{(j)}$, i.e., the trajectories over the next control horizon of human pedestrians by the Social-LSTM model \eqref{eq:lstm_recursive} where the inputs utilize the robots' trajectories from the previous iteration.
The coordinator transmits $\bb{s}_{\HHH}^{(j)}$ to all the robots.
Next, each robot solves its local MPC problem given other robots' best-response trajectories in parallel and decides to accept or reject the new solution based on whether or not it can improve the cost function by at least $\epsilon$.
Then, each robot $i$ transmits the current best-response trajectories $\bb{x}_i^{(j)}$ to the coordinator and receives $\bb{x}_{-i}^{(j)}$.
The algorithm is terminated if the coordinator detects that the condition $C_{i,t} (\bb{u}_{i}^{(j)}, \bb{u}_{-i}^{(j)}) \le C_{i,t} (\bb{u}_{i}^{(j-1)}, \bb{u}_{-i}^{(j)}) + \epsilon$ is true $\forall i \in \RRR $, or if $j = j_{\max}$.

\begin{algorithm}[tb!]
\small
  \caption{IBR-based distributed MPC}
  \label{alg:ibrdmpc}
  \begin{algorithmic}[1]
    \Require $t$, $H$, $j_{\mathrm{max}}$, $\epsilon$, $\bb{u}_{\RRR}^{(0)}$, $\bb{x}_{\RRR}^{(0)}$, $\bb{s}_{\RRR\HHH,t-L+1:t}$.
    
    \For{$j = 1,\dots,j_{\mathrm{max}}$}
    \State Coordinator predicts {$\bb{s}_{\HHH}^{(j)}$} by \eqref{eq:lstm_recursive} and transmits to $\RRR$.
    \For{robot $i \in \RRR$} {\textsc{(In parallel)}}
    \State Solve \eqref{eq:agent_mpc} given $\bb{s}_{\HHH}^{(j)}$ and $\bb{x}_{-i}^{(j-1)}$ to obtain $\bb{u}_{i}^{(j)}$ and $\bb{x}_{i}^{(j)}$.
    \If {$C_{i,t} (\bb{u}_{i}^{(j)}, \bb{u}_{-i}^{(j-1)}) \!>\! C_{i,t} (\bb{u}_{i}^{(j-1)}, \bb{u}_{-i}^{(j-1)}) - \epsilon$}
    \State $\bb{u}_{i}^{(j)} \leftarrow \bb{u}_{i}^{(j-1)}$
    \EndIf
    \State Compute and transmit $\bb{x}_{i}^{(j)}$ to the coordinator.
    \EndFor
    \State Coordinator receives $\bb{x}_{i}^{(j)}$ from $\RRR$, then transmits $\bb{x}_{-i}^{(j)}$ to each robot $i \in \RRR$.
    \For{robot $i \in \RRR$}  {\textsc{(In parallel)}} 
    \If {$C_{i,t} (\bb{u}_{i}^{(j)}, \bb{u}_{-i}^{(j)}) \le C_{i,t} (\bb{u}_{i}^{(j-1)}, \bb{u}_{-i}^{(j)}) + \epsilon$}
    \State $ \texttt{Conv}_i \leftarrow \texttt{True}$
    \Else 
    \State $ \texttt{Conv}_i \leftarrow \texttt{False}$
    \EndIf
    \State Transmit $\texttt{Conv}_i$ to the coordinator.
    \EndFor
    \If {Coordinator detects $\texttt{Conv}_i = \texttt{True}$, $\forall i \in \RRR$}
    \State \Return $\bb{u}_{\RRR}^{(j)}$ 
    \EndIf
    \EndFor

    \State \Return $\bb{u}_{\RRR}^{(j_{\mathrm{max}})}$ 
  \end{algorithmic}
\end{algorithm}  
\setlength{\textfloatsep}{0.05cm}

\section{Simulation Results}
\label{sec:sim}

\subsection{Simulation Setup}

We conducted simulated crowd navigation experiments in Python with the CrowdNav environment 
\cite{chen2019crowd}, wherein human pedestrians are simulated using ORCA \cite{van2008reciprocal}.
We utilized Trajnet++ benchmark 
for training Social-LSTM models on the ETH dataset \cite{pellegrini2009you}.
CasADi \cite{andersson2019casadi} and IPOPT solver \cite{wachter2006implementation} were used for formulating and solving the nonlinear MPC problems, respectively.
The framework parameters were chosen as given in Table~\ref{tab:parameters}.
The simulations were executed on an MSI computer with an Intel Core i9 CPU, \SI{64}{GB} RAM, and a GeForce RTX 3080 Ti GPU. 
Some of the results can be visualized online\footnotemark. 

\footnotetext[1]{\url{https://sites.google.com/view/crowdmrn}}

\begin{table}[h!]
\caption{Parameters of the MPC problem.}
\label{tab:parameters} 
\centering
\begin{tabular}{ p{0.09\textwidth}  p{0.10\textwidth} p{0.09\textwidth} p{0.10\textwidth} }
\toprule[1pt]
\textbf{Parameters} & \textbf{Values} & \textbf{Parameters} & \textbf{Values} \\
\midrule[0.5pt] 
$H$ & $4$ & $L$ & $8$ \\
$\tau$ & $\SI{0.4}{s}$ & $\omega_i^{\mathrm{goal}}$ & $10.0$ \\
$\omega_i^{\mathrm{acce}}$ & $10^{-1}$ &
$\omega_i^{\mathrm{jerk}}$ & $10^{-1}$ \\ $\omega_i^{\mathrm{coll}}$ & $10^{7}$ &
$\omega_i^{\mathrm{floc}}$ & $10$ \\ 
$v_\mathrm{max}$ & $\SI{1.0} {m/s}$ & $a_\mathrm{max}$ & $\SI{2.0} {m/s^2}$ \\
$\rho$ & $\SI{0.5} {s}$ & $d_\mathrm{min}$ & $\SI{0.8}{m}$ \\ 
$\mu$ & $30$ &
$j_{\max}$ & $10$ \\
$\xi$ & $10^{-3}$ & $\epsilon$ & $10^{-3}$  \\
\bottomrule[1pt] 
\end{tabular}
\end{table}

\subsection{Results and Discussions}
The trajectories of the robots and human pedestrians in a circular crossing simulation with 3 robots and 7 pedestrians are shown in Fig.~\ref{fig:trajcircle}.
As can be seen, due to the flocking objective, the robots move from their origins to form a flock with other robots.
The flock of robots can navigate among the humans without any collisions.
The robots depart from the flock when they are close to their goals, and all the robots eventually reach their goals within \SI{16.4}{s}.

To further assess the effectiveness of the proposed framework,
we analysed performance on the following metrics: 
\begin{itemize}
\item \textbf{Success rate}: \% of simulations in which all agents reach their destinations.
\item \textbf{Average travel time}: Time($\rm{s}$) for all robots to reach their destinations (in simulations with success).
\item \textbf{Collision rate}:  \% of simulations that the minimum distance between the robots and the pedestrians is less than \SI{0.8}{m} (violation of personal space).
\item \textbf{Discomfort rate}: \% of simulations wherein a robot’s projected path intersects with a human’s projected path. 
The projected path is defined as a line segment from the current agent's position along with the direction of the agent’s velocity and the length proportional to speed. 
The proportional factor is chosen as \SI{1.2}{s}. 
\end{itemize}
The success rate and average travel time quantify the efficiency, while the collision rate and discomfort rate are related to social conformity \cite{wang2022metrics} of the navigation algorithms.
Those validation metrics are computed by averaging $1000$ simulations with randomized initial conditions, including $500$ circular crossing simulations and $500$ perpendicular crossing simulations \cite{wang2022metrics}. 
We compared the metrics for centralized MPC and distributed MPC in Table~\ref{tab:cenvsdis} with different numbers of human pedestrians.
First, it can be observed that the performance deteriorates as the crowd density increases.
The centralized and distributed MPC algorithms perform similarly, with the centralized approach taking a narrow lead across most metrics.

Furthermore, we evaluated the benefits of flocking control in reducing personal space violations and discomfort to humans. 
The results for CMPC with and without the flocking control objective are tallied in Table~\ref{tab:flockvsno}.
The collision and discomfort rates can be reduced by including the flocking objective. Still, the robots generally take longer to reach their goals and decrease the number of successful simulations.
The results indicate that moving in a flock can indirectly improve social conformity, but there is a higher chance of ``freezing robots'' when the flock of robots cannot find a safe yet efficient trajectory to avoid deadlock.
This suggests a high-level reference trajectory generator can be combined with the proposed framework to avoid such situations.

\section{Conlusions}
\label{sec:conclu}

In this paper, we addressed the problem of cooperative navigation for multiple robots in crowds by combining game-theoretic MPC and a Social-LSTM human trajectory prediction model.
We employed an iterative best-response approach to develop two algorithms for solving the MPC problem in centralized and distributed manners.
The simulation results demonstrated the effectiveness of the control framework.
Potential directions for future research are to solve the MPC problem with fast-distributed optimization algorithms, investigate the effects of shared objectives, and validate the control framework through experiments.

\bibliographystyle{IEEEtran.bst} 
\bibliography{reference/ref.bib}

\begin{thebibliography}{10}
\providecommand{\url}[1]{#1}
\csname url@rmstyle\endcsname
\providecommand{\newblock}{\relax}
\providecommand{\bibinfo}[2]{#2}
\providecommand\BIBentrySTDinterwordspacing{\spaceskip=0pt\relax}
\providecommand\BIBentryALTinterwordstretchfactor{4}
\providecommand\BIBentryALTinterwordspacing{\spaceskip=\fontdimen2\font plus
\BIBentryALTinterwordstretchfactor\fontdimen3\font minus
  \fontdimen4\font\relax}
\providecommand\BIBforeignlanguage[2]{{%
\expandafter\ifx\csname l@#1\endcsname\relax
\typeout{** WARNING: IEEEtran.bst: No hyphenation pattern has been}%
\typeout{** loaded for the language `#1'. Using the pattern for}%
\typeout{** the default language instead.}%
\else
\language=\csname l@#1\endcsname
\fi
#2}}

\bibitem{mavrogiannis2023core}
C.~Mavrogiannis, F.~Baldini, A.~Wang, D.~Zhao, P.~Trautman, A.~Steinfeld, and
  J.~Oh, ``Core challenges of social robot navigation: A survey,'' \emph{ACM
  Transactions on Human-Robot Interaction}, vol.~12, no.~3, pp. 1--39, 2023.

\bibitem{Malikopoulos2021}
A.~A. Malikopoulos, ``On team decision problems with nonclassical information
  structures,'' \emph{IEEE Transactions on Automatic Control}, vol.~68, no.~7,
  pp. 3915--3930, 2023.

\bibitem{chalaki2023minimally}
B.~Chalaki, V.~Tadiparthi, H.~N. Mahjoub, J.~D'sa, E.~Moradi-Pari, A.~S.~C.
  Armijos, A.~Li, and C.~G. Cassandras, ``Minimally disruptive cooperative
  lane-change maneuvers,'' \emph{IEEE Control Systems Letters}, 2023.

\bibitem{su2021toward}
H.~Su, A.~Mariani, S.~E. Ovur, A.~Menciassi, G.~Ferrigno, and E.~De~Momi,
  ``Toward teaching by demonstration for robot-assisted minimally invasive
  surgery,'' \emph{IEEE Transactions on Automation Science and Engineering},
  vol.~18, no.~2, pp. 484--494, 2021.

\bibitem{alleyne2023control}
A.~Alleyne, F.~Allg{\"o}wer, A.~Ames, S.~Amin, J.~Anderson, A.~Annaswamy,
  P.~Antsaklis, N.~Bagheri, H.~Balakrishnan, B.~Bamieh, \emph{et~al.},
  ``Control for societal-scale challenges: Road map 2030,'' in \emph{2022 IEEE
  CSS Workshop on Control for Societal-Scale Challenges}.\hskip 1em plus 0.5em
  minus 0.4em\relax IEEE Control Systems Society, 2023.

\bibitem{chen2017socially}
Y.~F. Chen, M.~Everett, M.~Liu, and J.~P. How, ``Socially aware motion planning
  with deep reinforcement learning,'' in \emph{2017 IEEE/RSJ International
  Conference on Intelligent Robots and Systems (IROS)}.\hskip 1em plus 0.5em
  minus 0.4em\relax IEEE, 2017, pp. 1343--1350.

\bibitem{everett2018motion}
M.~Everett, Y.~F. Chen, and J.~P. How, ``Motion planning among dynamic,
  decision-making agents with deep reinforcement learning,'' in \emph{2018
  IEEE/RSJ International Conference on Intelligent Robots and Systems
  (IROS)}.\hskip 1em plus 0.5em minus 0.4em\relax IEEE, 2018, pp. 3052--3059.

\bibitem{chen2019crowd}
C.~Chen, Y.~Liu, S.~Kreiss, and A.~Alahi, ``Crowd-robot interaction:
  Crowd-aware robot navigation with attention-based deep reinforcement
  learning,'' in \emph{2019 international conference on robotics and automation
  (ICRA)}.\hskip 1em plus 0.5em minus 0.4em\relax IEEE, 2019, pp. 6015--6022.

\bibitem{liu2023intention}
S.~Liu, P.~Chang, Z.~Huang, N.~Chakraborty, K.~Hong, W.~Liang, D.~L. McPherson,
  J.~Geng, and K.~Driggs-Campbell, ``Intention aware robot crowd navigation
  with attention-based interaction graph,'' in \emph{2023 IEEE International
  Conference on Robotics and Automation (ICRA)}.\hskip 1em plus 0.5em minus
  0.4em\relax IEEE, 2023, pp. 12\,015--12\,021.

\bibitem{brito2021go}
B.~Brito, M.~Everett, J.~P. How, and J.~Alonso-Mora, ``Where to go next:
  Learning a subgoal recommendation policy for navigation in dynamic
  environments,'' \emph{IEEE Robotics and Automation Letters}, vol.~6, no.~3,
  pp. 4616--4623, 2021.

\bibitem{akhtyamov2023social}
T.~Akhtyamov, A.~Kashirin, A.~Postnikov, and G.~Ferrer, ``Social robot
  navigation through constrained optimization: a comparative study of
  uncertainty-based objectives and constraints,'' \emph{arXiv preprint
  arXiv:2305.02859}, 2023.

\bibitem{vulcano2022safe}
V.~Vulcano, S.~G. Tarantos, P.~Ferrari, and G.~Oriolo, ``Safe robot navigation
  in a crowd combining nmpc and control barrier functions,'' in \emph{2022 IEEE
  61st Conference on Decision and Control (CDC)}.\hskip 1em plus 0.5em minus
  0.4em\relax IEEE, 2022, pp. 3321--3328.

\bibitem{chen2021interactive}
Y.~Chen, F.~Zhao, and Y.~Lou, ``Interactive model predictive control for robot
  navigation in dense crowds,'' \emph{IEEE Transactions on Systems, Man, and
  Cybernetics: Systems}, vol.~52, no.~4, pp. 2289--2301, 2021.

\bibitem{poddar2023crowd}
S.~Poddar, C.~Mavrogiannis, and S.~S. Srinivasa, ``From crowd motion prediction
  to robot navigation in crowds,'' \emph{arXiv preprint arXiv:2303.01424},
  2023.

\bibitem{lindemann2023safe}
L.~Lindemann, M.~Cleaveland, G.~Shim, and G.~J. Pappas, ``Safe planning in
  dynamic environments using conformal prediction,'' \emph{IEEE Robotics and
  Automation Letters}, 2023.

\bibitem{alahi2016social}
A.~Alahi, K.~Goel, V.~Ramanathan, A.~Robicquet, L.~Fei-Fei, and S.~Savarese,
  ``Social lstm: Human trajectory prediction in crowded spaces,'' in
  \emph{Proceedings of the IEEE conference on computer vision and pattern
  recognition}, 2016, pp. 961--971.

\bibitem{gupta2018social}
A.~Gupta, J.~Johnson, L.~Fei-Fei, S.~Savarese, and A.~Alahi, ``Social gan:
  Socially acceptable trajectories with generative adversarial networks,'' in
  \emph{Proceedings of the IEEE conference on computer vision and pattern
  recognition}, 2018, pp. 2255--2264.

\bibitem{liu2021social}
Y.~Liu, Q.~Yan, and A.~Alahi, ``Social nce: Contrastive learning of
  socially-aware motion representations,'' in \emph{Proceedings of the IEEE/CVF
  International Conference on Computer Vision}, 2021, pp. 15\,118--15\,129.

\bibitem{trautman2017sparse}
P.~Trautman, ``Sparse interacting gaussian processes: Efficiency and optimality
  theorems of autonomous crowd navigation,'' in \emph{2017 IEEE 56th Annual
  Conference on Decision and Control (CDC)}.\hskip 1em plus 0.5em minus
  0.4em\relax IEEE, 2017, pp. 327--334.

\bibitem{zhu2021learning}
H.~Zhu, F.~M. Claramunt, B.~Brito, and J.~Alonso-Mora, ``Learning
  interaction-aware trajectory predictions for decentralized multi-robot motion
  planning in dynamic environments,'' \emph{IEEE Robotics and Automation
  Letters}, vol.~6, no.~2, pp. 2256--2263, 2021.

\bibitem{la2016potential}
Q.~D. L{\~a}, Y.~H. Chew, and B.-H. Soong, ``Potential game theory,''
  \emph{Cham: Springer International Publishing}, 2016.

\bibitem{gupta2023interaction}
P.~Gupta, D.~Isele, D.~Lee, and S.~Bae, ``Interaction-aware trajectory planning
  for autonomous vehicles with analytic integration of neural networks into
  model predictive control,'' \emph{arXiv preprint arXiv:2301.05393}, 2023.

\bibitem{kothari2021human}
P.~Kothari, S.~Kreiss, and A.~Alahi, ``Human trajectory forecasting in crowds:
  A deep learning perspective,'' \emph{IEEE Transactions on Intelligent
  Transportation Systems}, vol.~23, no.~7, pp. 7386--7400, 2021.

\bibitem{wang2020multi}
Z.~Wang, T.~Taubner, and M.~Schwager, ``Multi-agent sensitivity enhanced
  iterative best response: A real-time game theoretic planner for drone racing
  in 3d environments,'' \emph{Robotics and Autonomous Systems}, vol. 125, p.
  103410, 2020.

\bibitem{fisac2019hierarchical}
J.~F. Fisac, E.~Bronstein, E.~Stefansson, D.~Sadigh, S.~S. Sastry, and A.~D.
  Dragan, ``Hierarchical game-theoretic planning for autonomous vehicles,'' in
  \emph{2019 International conference on robotics and automation (ICRA)}.\hskip
  1em plus 0.5em minus 0.4em\relax IEEE, 2019, pp. 9590--9596.

\bibitem{liu2023potential}
M.~Liu, I.~Kolmanovsky, H.~E. Tseng, S.~Huang, D.~Filev, and A.~Girard,
  ``Potential game-based decision-making for autonomous driving,'' \emph{IEEE
  Transactions on Intelligent Transportation Systems}, 2023.

\bibitem{williams2018best}
G.~Williams, B.~Goldfain, P.~Drews, J.~M. Rehg, and E.~A. Theodorou, ``Best
  response model predictive control for agile interactions between autonomous
  ground vehicles,'' in \emph{2018 IEEE International Conference on Robotics
  and Automation (ICRA)}.\hskip 1em plus 0.5em minus 0.4em\relax IEEE, 2018,
  pp. 2403--2410.

\bibitem{espinoza2022deep}
J.~L.~V. Espinoza, A.~Liniger, W.~Schwarting, D.~Rus, and L.~Van~Gool, ``Deep
  interactive motion prediction and planning: Playing games with motion
  prediction models,'' in \emph{Learning for Dynamics and Control
  Conference}.\hskip 1em plus 0.5em minus 0.4em\relax PMLR, 2022, pp.
  1006--1019.

\bibitem{roughgarden2010algorithmic}
T.~Roughgarden, ``Algorithmic game theory,'' \emph{Communications of the ACM},
  vol.~53, no.~7, pp. 78--86, 2010.

\bibitem{van2008reciprocal}
J.~Van~den Berg, M.~Lin, and D.~Manocha, ``Reciprocal velocity obstacles for
  real-time multi-agent navigation,'' in \emph{2008 IEEE international
  conference on robotics and automation}.\hskip 1em plus 0.5em minus
  0.4em\relax Ieee, 2008, pp. 1928--1935.

\bibitem{pellegrini2009you}
S.~Pellegrini, A.~Ess, K.~Schindler, and L.~Van~Gool, ``You'll never walk
  alone: Modeling social behavior for multi-target tracking,'' in \emph{2009
  IEEE 12th international conference on computer vision}.\hskip 1em plus 0.5em
  minus 0.4em\relax IEEE, 2009, pp. 261--268.

\bibitem{andersson2019casadi}
J.~A. Andersson, J.~Gillis, G.~Horn, J.~B. Rawlings, and M.~Diehl, ``Casadi: a
  software framework for nonlinear optimization and optimal control,''
  \emph{Mathematical Programming Computation}, vol.~11, pp. 1--36, 2019.

\bibitem{wachter2006implementation}
A.~W{\"a}chter and L.~T. Biegler, ``On the implementation of an interior-point
  filter line-search algorithm for large-scale nonlinear programming,''
  \emph{Mathematical programming}, vol. 106, pp. 25--57, 2006.

\bibitem{wang2022metrics}
J.~Wang, W.~P. Chan, P.~Carreno-Medrano, A.~Cosgun, and E.~Croft, ``Metrics for
  evaluating social conformity of crowd navigation algorithms,'' in \emph{2022
  IEEE International Conference on Advanced Robotics and Its Social Impacts
  (ARSO)}.\hskip 1em plus 0.5em minus 0.4em\relax IEEE, 2022, pp. 1--6.

\end{thebibliography}

\clearpage
\newpage
\end{document}